\documentclass[11pt,a4paper]{article}
\usepackage[hyperref]{naaclhlt2019}
\usepackage{times}
\usepackage{latexsym}

\usepackage{url}
\usepackage{graphicx}
\usepackage{hhline}
\usepackage{tcolorbox}
\usepackage{color}
\usepackage{amsmath}
\usepackage{amsfonts}
\usepackage{booktabs}
\usepackage{multirow}
\usepackage{makecell}
\usepackage{enumitem}

\usepackage[linesnumbered,algoruled,boxed,lined,noend]{algorithm2e}

\newcolumntype{C}[1]{>{\centering\arraybackslash}m{#1}}

\DeclareMathOperator*{\argmax}{argmax} 

\graphicspath{{./figs/}}

\aclfinalcopy 
\def\aclpaperid{181} 

\title{Integrating Semantic Knowledge to Tackle Zero-shot Text Classification}

\author{Jingqing Zhang \thanks{~~~Piyawat Lertvittayakumjorn and Jingqing Zhang contributed equally to this project. } \\
  Data Science Institute \\
  Imperial College London \\ 
  London, UK\\ \And
  Piyawat Lertvittayakumjorn \footnotemark[1] \\
  Department of Computing \\
  Imperial College London \\
  London, UK \\
  {\tt \{jingqing.zhang15,pl1515,y.guo\}@imperial.ac.uk} \\ \And
  Yike Guo\\
  Data Science Institute \\
  Imperial College London \\ 
  London, UK
}

\date{}

\begin{document}
\maketitle
\begin{abstract}
Insufficient or even unavailable training data of emerging classes is a big challenge of many classification tasks, including text classification.
Recognising text documents of classes that have never been seen in the learning stage, so-called \textit{zero-shot text classification}, is therefore difficult and only limited previous works tackled this problem.
In this paper, we propose a two-phase framework together with data augmentation and feature augmentation to solve this problem. Four kinds of semantic knowledge (word embeddings, class descriptions, class hierarchy, and a general knowledge graph) are incorporated into the proposed framework to deal with instances of unseen classes effectively. Experimental results show that each and the combination of the two phases achieve the best overall accuracy compared with baselines and recent approaches in classifying real-world texts under the zero-shot scenario. 
\end{abstract}


\begin{figure*}[!htb]
\centering
\includegraphics[width=\textwidth]{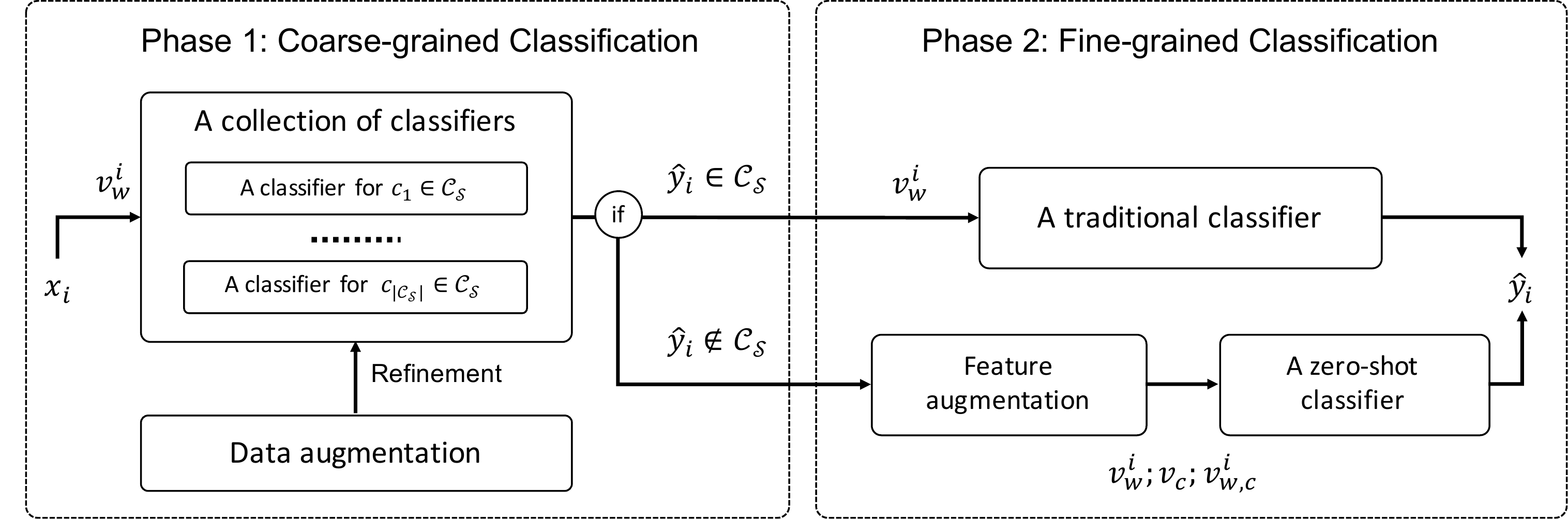}
\caption{The overview of the proposed framework with two phases. The coarse-grained phase judges if an input document $x_i$ comes from seen or unseen classes. The fine-grained phase finally decides the class $\hat{y}_i$. All notations are defined in section \ref{subsec:problem}-\ref{subsec:overview_notations}.}\label{fig:framework}
\end{figure*}

\section{Introduction} \label{sec:int}
As one of the most fundamental problems in machine learning, automatic classification has been widely studied in several domains. However, many approaches, proven to be effective in traditional classification tasks, cannot catch up with a dynamic and open environment where new classes can emerge after the learning stage \cite{2015-Romera-EmbarrassinglyZSL}. For example, the number of topics on social media is growing rapidly, and the classification models are required to recognise the text of the new topics using only general information (e.g., descriptions of the topics) since labelled training instances are unfeasible to obtain for each new topic \cite{lee2011twitter}. This scenario holds in many real-world domains such as object recognition and medical diagnosis \cite{2017-Xian-ZSLGoodBadUgly,world1996infectious}. 

Zero-shot learning (ZSL) for text classification aims to classify documents of classes which are absent from the learning stage. 
Although it is challenging for a machine to achieve, 
humans are able to learn new concepts by transferring knowledge from known to unknown domains based on high-level descriptions and semantic representations \cite{thrun1998learning}.
Therefore, without labelled data of unseen classes, a zero-shot learning framework is expected to exploit supportive semantic knowledge (e.g., class descriptions, relations among classes, and external domain knowledge) to generally infer the features of unseen classes using patterns learned from seen classes. 

So far, three main types of semantic knowledge have been employed in general zero-shot scenarios \cite{2017-Fu-RecentAdvancesZeroShot}. 
The most widely used one is semantic attributes of classes such as visual concepts (e.g., colours, shapes) and semantic properties (e.g., behaviours, functions) \cite{2009-Lampert-ZSLAttributes,2018-Zhao-LAD}.
The second type is concept ontology, including class hierarchy and knowledge graphs, which represents relationships among classes and features 
\cite{2018-Wang-ZSRwithEmbKG,2010-Fergus-LabelSharing}.
The third type is semantic word embeddings which capture implicit relationships between words thanks to a large training text corpus  
 \cite{2013-Socher-ZSLCrossModal,2013-Norouzi-zero}.
Nonetheless, concerning ZSL in text classification particularly, there are few studies exploiting one of these knowledge types and none has considered the combinations of them \cite{2017-Pushp-TOTA,2013-Dauphin-ZSLSemanticUtterance}.  
Moreover, some previous works used different datasets to train and test, 
but there is similarity between classes in the training and testing set. For example, in \cite{2013-Dauphin-ZSLSemanticUtterance}, the class ``imdb.com'' in the training set naturally corresponds to the class ``Movies'' in the testing set. Hence, these methods are not working under a strict zero-shot scenario.

To tackle the zero-shot text classification problem, this paper proposes a novel two-phase framework together with data augmentation and feature augmentation (Figure \ref{fig:framework}). In addition, four kinds of semantic knowledge including word embeddings, class descriptions, class hierarchy, and a general knowledge graph (ConceptNet) are exploited in the framework to effectively learn the unseen classes. 
Both of the two phases are based on convolutional neural networks \cite{kim2014convolutional}. 
The first phase called \textbf{coarse-grained classification} judges if a document is from seen or unseen classes. 
Then, the second phase, named \textbf{fine-grained classification}, finally decides its class. 
Note that all the classifiers in this framework are trained using labelled data of seen classes (and augmented text data) only. None of the steps learns from the labelled data of unseen classes. 


The contributions of our work can be summarised as follows.

\begin{itemize}
    \item We propose a novel deep learning based two-phase framework, including coarse-grained and fine-grained classification, to tackle the zero-shot text classification problem. Unlike some previous works, our framework does not require semantic correspondence between classes in a training stage and classes in an inference stage. In other words, the seen and unseen classes can be clearly different.
    \item We propose a novel data augmentation technique called topic translation to strengthen the capability of our framework to detect documents from unseen classes effectively.
    \item We propose a method to perform feature augmentation by using integrated semantic knowledge to transfer the knowledge learned from seen to unseen classes in the zero-shot scenario.
\end{itemize}

In the remainder of this paper, we firstly explain our proposed zero-shot text classification framework in section \ref{sec:met}. Experiments and results, which demonstrate the performance of our framework, are presented in section \ref{sec:exp}. Related works are discussed in section \ref{sec:rel}. Finally, section \ref{sec:con} concludes our work and mentions possible future work.

\section{Methodology}\label{sec:met}

\subsection{Problem Formulation} \label{subsec:problem}
\begin{figure}[!ht]
\centering
\includegraphics[width=0.48\textwidth]{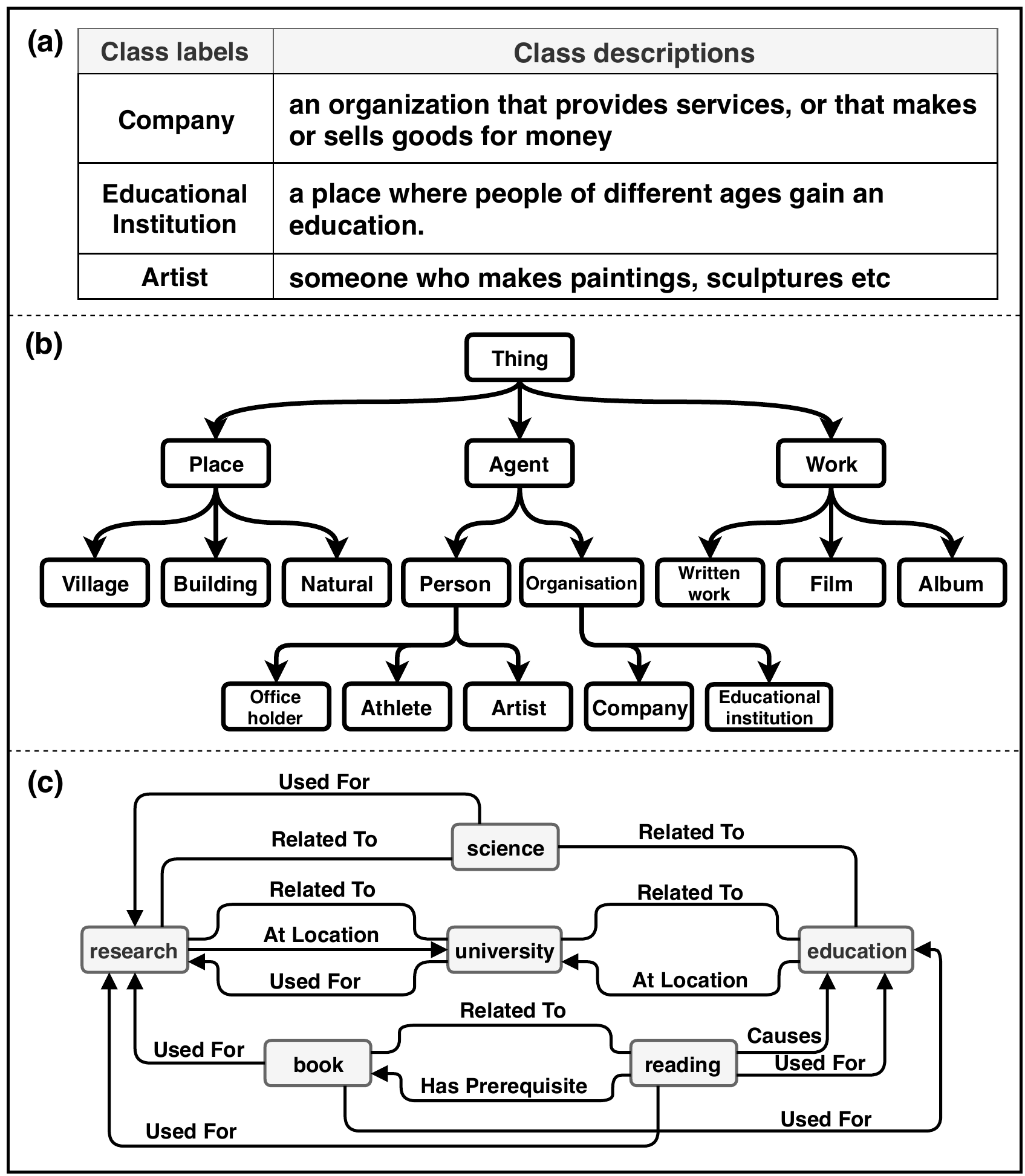}
\caption{Illustrations of semantic knowledge integrated into our framework: (a) class labels and class descriptions (b) class hierarchy and (c) a subgraph of the general knowledge graph (ConceptNet).}\label{fig:semantic}
\end{figure}

Let $\mathcal{C_S}$ and $\mathcal{C_U}$ be disjoint sets of seen and unseen classes of the classification respectively.
In the learning stage, a training set $\{ (x_1, y_1), \ldots, (x_n, y_n) \}$ is given where $x_i$ is the $i$-th document containing a sequence of words $[w_1^i,w_2^i,\dots,w_t^i]$ and $y_i \in \mathcal{C_S}$ is the class of $x_i$. In the inference stage, the goal is to predict the class of each document, $\hat{y_i}$, in a testing set which has the same data format as the training set except that $y_i$ comes from $\mathcal{C_S} \cup \mathcal{C_U}$. Note that \textit{(i)} every class comes with a class label and a class description (Figure \ref{fig:semantic}a); \textit{(ii)} a class hierarchy showing superclass-subclass relationships is also provided (Figure \ref{fig:semantic}b); \textit{(iii)} the documents from unseen classes cannot be observed to train the framework. 

\subsection{Overview and Notations}
\label{subsec:overview_notations}
As discussed in the Introduction, our proposed classification framework consists of two phases (Figure \ref{fig:framework}). 
The first phase, coarse-grained classification, predicts whether an input document comes from seen or unseen classes. We also apply a data augmentation technique in this phase to help the classifiers be aware of the existence of unseen classes without accessing their real data. 
Then the second phase, fine-grained classification, finally specifies the class of the input document. It uses either a traditional classifier or a zero-shot classifier depending on the coarse-grained prediction given by Phase 1. 
Also, feature augmentation based on semantic knowledge is used to provide additional information which relates the document and the unseen classes to generalise the zero-shot reasoning.

We use the following notations in Figure \ref{fig:framework} and throughout this paper.
\begin{itemize}
    \item The list of embeddings of each word in the document $x_i$ is denoted by $v_w^i=[v_{w_1}^i, v_{w_2}^i, \dots, v_{w_t}^i]$.
    \item The embedding of each class label $c$ is denoted by $v_c$, $\forall c \in \mathcal{C_S}\cup\mathcal{C_U}$. It is assumed that each class has a one-word class label. If the class label has more than one word, a similar one-word class label is provided to find $v_c$.
    \item As augmented features, the relationship vector $v_{w_j,c}^i$ shows the degree of relatedness between the word $w_j$ and the class $c$ according to semantic knowledge. Hence, the list of relationship vectors between each word in $x_i$ and each class $c \in \mathcal{C_S}\cup\mathcal{C_U}$ is denoted by $v_{w,c}^i$ $=[v_{w_1,c}^i, v_{w_2,c}^i, \dots, v_{w_t,c}^i]$. We will explain the construction method in section \ref{subsec:feature_aug}.
\end{itemize}

\subsection{Phase 1: Coarse-grained Classification} \label{subsec:phase1}

Given a document $x_i$, Phase 1 performs a binary classification to decide whether $\hat{y}_i\in\mathcal{C_S}$ or $\hat{y}_i\notin\mathcal{C_S}$.
In this phase, each seen class $c_s \in \mathcal{C_S}$ has its own CNN classifier (with a subsequent dense layer and a sigmoid output) to predict the confidence that $x_i$ comes from the class $c_s$, i.e., $p(\hat{y}_i=c_s|x_i)$. 
The classifier uses $v_w^i$ as an input and it is trained using a binary cross entropy loss with all documents of its class in the training set as positive examples and the rest as negative examples. 

For a test document $x_i$, this phase computes $p(\hat{y}_i=c_s|x_i)$ for every seen class $c_s$ in $\mathcal{C_S}$. If there exists a class $c_s$ such that $p(\hat{y}_i=c_s|x_i) > \tau_s$, it predicts $\hat{y}_i \in \mathcal{C_S}$; otherwise, $\hat{y}_i \notin \mathcal{C_S}$. $\tau_s$ is a classification threshold for the class $c_s$, calculated based on the threshold adaptation method from \cite{2017-Shu-DOC}.



\subsubsection{Data Augmentation}
During the learning stage, the classifiers in Phase 1 use negative examples solely from seen classes, so they may not be able to differentiate the positive class from unseen classes. Hence, when the names of unseen classes are known in the inference stage, we try to introduce them to the classifiers in Phase 1 via augmented data so they can learn to reject the instances likely from unseen classes. We do data augmentation by translating a document from its original seen class to a new unseen class using analogy. We call this process \textit{topic translation}. 

In the word level, we translate a word $w$ in a document of class $c$ to a corresponding word $w'$ in the context of a target class $c'$ by solving an analogy question ``$c$:$w$ :: $c'$:?''. For example, solving the analogy ``company:firm :: village:?'' via word embeddings \cite{2013-Mikolov-Word2Vec}, we know that the word ``firm'' in a document of class ``company'' can be translated into the word ``hamlet'' in the context of class ``village''. Our framework adopts the \textsc{3CosMul} method by \citet{levy2014linguistic} to solve the analogy question and find candidates of $w'$: 
\[w' = \underset{x \in V}\argmax \frac{\cos(x, c')\cos(x, w)}{\cos(x, c) + \epsilon}\]
where $V$ is a vocabulary set and $\cos(a,b)$ is a cosine similarity score between the vectors of word $a$ and word $b$. Also, $\epsilon$ is a small number (i.e., 0.001) added to prevent division by zero.

In the document level, we follow Algorithm \ref{alg:topictransfer} to translate a document of class $c$ into the topic of another class $c'$. To explain, we translate all nouns, verbs, adjectives, and adverbs in the given document to the target class, word-by-word, using the word-level analogy. The word to replace must have the same part of speech as the original word and all the replacements in one document are 1-to-1 relations, enforced by replace\_dict in Algorithm \ref{alg:topictransfer}. With this idea, we can create augmented documents for the unseen classes by topic-translation from the documents of seen classes in the training dataset. After that, we can use the augmented documents as additional negative examples for all the CNNs in Phase 1 to make them aware of the tone of unseen classes.

\begin{algorithm}[ht]
\small
\SetAlgoLined
\SetKwInput{Precondition}{Precondition~}
\SetKwInput{Input}{Input~}
\Input{a document $x_i = [w_1^i,w_2^i,\dots,w_t^i]$, \\\quad\qquad an original class label $c$, a target class label $c'$ 
}
\KwOut{a translated document $x_i'$}
replace\_dict = \textit{dict}(); $x_i'$ = []\;
\ForEach{$w \in x_i$}{
    \eIf{\upshape is\_valid\_pos($w$)}{
        \If{$w \notin$ \upshape replace\_dict}{
            cands = solve\_analogy($w$, $c$, $c'$, top\_k=20)\;
            \For{\upshape$j$ = 0 \KwTo len(cands)-1}{
                \If{\upshape cands[$j$] $\notin$ replace\_dict.values() $\wedge$ pos\_of($w$) $\in$ pos\_list(cands[$j$])}{
                    replace\_dict[$w$] = cands[$j$]\;
                    \textbf{break}\;
                }
            }
            \If{\upshape $j$ == len(cands)}{
                $x_i'$.append($w$)\;
                \textbf{continue}\;
            }
        }
        $x_i'$.append(replace\_dict[$w$])\;
   }{
   $x_i'$.append($w$)\;
   }
}
\KwRet{$x_i'$}
\caption{\small{Document-level topic translation}}
\label{alg:topictransfer}
\end{algorithm}


\subsection{Phase 2: Fine-grained Classification} \label{subsec:phase2}
Phase 2 decides the most appropriate class $\hat{y}_i$ for $x_i$ using two CNN classifiers: a traditional classifier and a zero-shot classifier as shown in Figure \ref{fig:framework}. If $\hat{y}_i \in \mathcal{C_S}$ predicted by Phase 1, the traditional classifier will finally select a class $c_s \in \mathcal{C_S}$ as $\hat{y}_i$. Otherwise, if $\hat{y}_i \notin \mathcal{C_S}$, the zero-shot classifier will be used to select a class $c_u \in \mathcal{C_U}$ as $\hat{y}_i$.

The traditional classifier and the zero-shot classifier have an identical CNN-based structure followed by two dense layers but their inputs and outputs are different. 
The traditional classifier is a multi-class classifier ($|\mathcal{C_S}|$ classes) with a softmax output, so it requires only the word embeddings $v_w^i$ as an input. This classifier is trained using a cross entropy loss with a training dataset whose examples are from seen classes only. 

In contrast, the zero-shot classifier is a binary classifier with a sigmoid output. Specifically, it takes a text document $x_i$ and a class $c$ as inputs and predicts the confidence $p(\hat{y}_i=c|x_i)$. However, in practice, we utilise $v_w^i$ to represent $x_i$, $v_c$ to represent the class $c$, and also augmented features $v_{w,c}^i$ to provide more information on how intimate the connections between words and the class $c$ are. Altogether, for each word $w_j$, the classifier receives the concatenation of three vectors (i.e., $[v_{w_j}^i; v_c; v_{w_j, c}^i]$) as an input. This classifier is trained using a binary cross entropy loss with a training data from seen classes only, but we expect this classifier to work well on unseen classes thanks to the distinctive patterns of $v_{w,c}^i$ in positive examples of every class. This is how we transfer knowledge from seen to unseen classes in ZSL. 


\subsubsection{Feature Augmentation} \label{subsec:feature_aug}
The relationship vector $v_{w_j,c}$ contains augmented features we input to the zero-shot classifier. $v_{w_j,c}$ shows how the word $w_j$ and the class $c$ are related considering the relations in a general knowledge graph. In this work, we use ConceptNet providing general knowledge of natural language words and phrases \cite{2013-Speer-ConceptNet}. A subgraph of ConceptNet is shown in Figure \ref{fig:semantic}c as an illustration. Nodes in ConceptNet are words or phrases, while edges connecting two nodes show how they are related either syntactically or semantically. 

We firstly represent a class $c$ as three sets of nodes in ConceptNet by processing the class hierarchy, class label, and class description of $c$. 
(1) \textit{\textbf{the\_class\_nodes}} is a set of nodes of the class label $c$ and any tokens inside $c$ if $c$ has more than one word.
(2) \textit{\textbf{superclass\_nodes}} is a set of nodes of all the superclasses of $c$ according to the class hierarchy. 
(3) \textit{\textbf{description\_nodes}} is a set of nodes of all nouns in the description of the class $c$. For example, if $c$ is the class ``Educational Institution'', according to Figure \ref{fig:semantic}a-\ref{fig:semantic}b, the three sets of ConceptNet nodes for this class are: 

\noindent(1) educational\_institution, educational, institution

\noindent(2) organization, agent

\noindent(3) place, people, ages, education.

To construct $v_{w_j,c}$, we consider whether the word $w_j$ is connected to the members of the three sets above within $K$ hops by particular types of relations or not\footnote{In this paper, we only consider the most common types of positive relations which are \textit{RelatedTo}, \textit{IsA}, \textit{PartOf}, and \textit{AtLocation}. They cover $\sim$60\% of all edges in ConceptNet.}. For each of the three sets, we construct a vector with $3K+1$ dimensions. 
\begin{itemize}
    \item $v[0] = 1$ if $w_j$ is a node in that set; otherwise, $v[0] = 0$.
    \item for $k=0, \dots, K-1$:
    \begin{itemize}
        \item $v[3k+1] = 1$ if there is a node in the set whose shortest path to $w_j$ is $k+1$. Otherwise, $v[3k+1] = 0$.
        \item $v[3k+2]$ is the number of nodes in the set whose shortest path to $w_j$ is $k+1$.
        \item $v[3k+3]$ is $v[3k+2]$ divided by the total number of nodes in the set.  
    \end{itemize}
\end{itemize} 
Thus, the vector associated to each set shows how $w_j$ is semantically close to that set. 
Finally, we concatenate the constructed vectors from the three sets to become $v_{w_j,c}$ with $3 \times (3K+1)$ dimensions. 




\section{Experiments} \label{sec:exp}

\subsection{Datasets}
We used two textual datasets for the experiments. The vocabulary size of each dataset was limited by 20,000 most frequent words and all numbers were excluded.
(1) \textbf{DBpedia} ontology dataset \cite{zhang2015character} includes 14 non-overlapping classes and textual data collected from Wikipedia. Each class has 40,000 training and 5,000 testing samples. 
(2) The \textbf{20newsgroups} dataset \footnote{http://qwone.com/$\sim$jason/20Newsgroups/} has 20 topics each of which has approximately 1,000 documents. 70\% of the documents of each class were randomly selected for training, and the remaining 30\% were used as a testing set.

\subsection{Implementation Details \footnote{Code: https://github.com/JingqingZ/KG4ZeroShotText.}}

In our experiments, two different rates of unseen classes, 50\% and 25\%, were chosen and the corresponding sizes of $\mathcal{C_S}$ and $\mathcal{C_U}$ are shown in Table \ref{tab:unseen_rate}.  For each dataset and each unseen rate, the random selection of ($\mathcal{C_S}$, $\mathcal{C_U}$) were repeated ten times and these ten groups were used by all the experiments with this setting for a fair comparison.
All documents from $\mathcal{C_U}$ were removed from the training set accordingly.
Finally, the results from all the ten groups were averaged.

In Phase 1, the structure of each classifier was identical. The CNN layer had three filter sizes [3, 4, 5] with 400 filters for each filter size and the subsequent dense layer had 300 units. For data augmentation, we used gensim with an implementation of \textsc{3CosMul} \cite{rehurek2010gensim} to solve the word-level analogy (line 5 in Algorithm \ref{alg:topictransfer}). Also, the numbers of augmented text documents per unseen class for every setting (if used) are indicated in Table \ref{tab:unseen_rate}. These numbers were set empirically considering the number of available training documents to be translated. 

In Phase 2, the traditional classifier and the zero-shot classifier had the same structure, in which the CNN layer had three filter sizes [2, 4, 8] with 600 filters for each filter size and the two intermediate dense layers had 400 and 100 units respectively.  
For feature augmentation, the maximum path length $K$ in ConceptNet was set to 3 to create the relationship vectors\footnote{Based on our observation, most of the related words stay within 3 hops from the class nodes in ConceptNet.}. 
The DBpedia ontology\footnote{http://mappings.dbpedia.org/server/ontology/classes/} was used to construct a class hierarchy of the DBpedia dataset. The class hierarchy of the 20newsgroups dataset was constructed based on the namespaces initially provided by the dataset. Meanwhile, the classes descriptions of both datasets were picked from Macmillan Dictionary\footnote{https://www.macmillandictionary.com/} as appropriate.

For both phases, we used 200-dim GloVe vectors\footnote{glove6B.zip in https://nlp.stanford.edu/projects/glove/} for word embeddings $v_w$ and $v_c$ \cite{pennington2014glove}. All the deep neural networks were implemented with TensorLayer \cite{dong2017tensorlayer} and TensorFlow \cite{martin2016tensorflow}.

\begin{table}[ht!]
\caption{The rates of unseen classes and the numbers of augmented documents (per unseen class) in the experiments}
\label{tab:unseen_rate}
\centering
\small
\begin{tabular}{| C{1.7cm} | C{1cm} | C{0.7cm} | C{0.7cm} | C{1.7cm} |} 
  \hline
  Dataset & Unseen rate & $|~\mathcal{C_S}~|$ & $|~\mathcal{C_U}~|$ & \#Augmented docs per $c_u$  \\
  \hline
  DBpedia & 25\% & 11 & 3 & 12,000 \\
  \cline{2-5}
  (14 classes) & 50\% & 7 & 7 & 8,000 \\ \hline
  20news & 25\% & 15 & 5 & 4,000 \\ 
  \cline{2-5}
  (20 classes) & 50\% & 10 & 10 & 3,000 \\
  \hline
\end{tabular}
\end{table}

\subsection{Baselines and Evaluation Metrics} \label{subsec:baselines}
We compared each phase and the overall framework with the following approaches and settings.

    \textbf{Phase 1:} Proposed by \cite{2017-Shu-DOC}, \textbf{DOC} is a state-of-the-art open-world text classification approach which classifies a new sample into a seen class or ``reject'' if the sample does not belong to any seen classes. The DOC uses a single CNN and a 1-vs-rest sigmoid output layer with threshold adjustment. Unlike DOC, the classifiers in the proposed Phase 1 work individually. However, for a fair comparison, we used DOC only as a binary classifier in this phase ($\hat{y}_i \in \mathcal{C_S}$ or $\hat{y}_i \notin \mathcal{C_S}$). 
    
    \textbf{Phase 2:} To see how well the augmented feature $v_{w,c}$ work in ZSL, we ran the zero-shot classifier with \textbf{different combinations of inputs}. Particularly, five combinations of $v_w$, $v_c$, and $v_{w,c}$ were tested with documents from unseen classes only (traditional ZSL).
    
    \textbf{The whole framework:}
    (1) \textbf{Count-based model} selected the class whose label appears most frequently in the document as $\hat{y}_i$. 
    (2) \textbf{Label similarity} \cite{sappadla2016using} is an unsupervised approach which calculates the cosine similarity between the sum of word embeddings of each class label and the sum of word embeddings of every n-gram ($n=1,2,3$) in the document. 
    We adopted this approach to do single-label classification by predicting the class that got the highest similarity score among all classes.
    (3) \textbf{RNN AutoEncoder} was built based on a Seq2Seq model 
    with LSTM (512 hidden units),
    and it was trained to encode documents and class labels onto the same latent space. The cosine similarity was applied to select a class label closest to the document on the latent space.  
    (4) \textbf{RNN+FC} refers to the architecture 2 proposed in \cite{2017-Pushp-TOTA}. It used an RNN layer with LSTM (512 hidden units) followed by two dense layers with 400 and 100 units respectively. 
    (5) \textbf{CNN+FC} replaced the RNN in the previous model with a CNN, which has the identical structure as the zero-shot classifier in Phase 2. Both RNN+FC and CNN+FC predicted the confidence $p(\hat{y}_i=c|x_i)$ given $v_w$ and $v_c$. The class with the highest confidence was selected as $\hat{y}_i$. 

For Phase 1, we used the accuracy for binary classification ($y, \hat{y}_i\in\mathcal{C_S}$ or $y, \hat{y}_i\notin\mathcal{C_S}$) as an evaluation metric. In contrast, for Phase 2 and the whole framework, we used the multi-class classification accuracy ($\hat{y}_i=y_i$) as a metric.

\begin{table*}[ht!]
 \caption{The accuracy of the whole framework compared with the baselines.} \label{tab:overall}
\centering
\small
\begin{tabular}{| c | c | c || C{0.1\linewidth} | C{0.1\linewidth} | C{0.1\linewidth} | C{0.1\linewidth} | C{0.1\linewidth} | C{0.06\linewidth} |} 
  \hline
  Dataset & \makecell{Unseen \\ rate} & $y_i$ & Count-based & Label Similarity \cite{sappadla2016using} & \makecell{RNN \\ Autoencoder} & RNN + FC \cite{2017-Pushp-TOTA} & CNN + FC & Ours \\
  \hhline{|=|=|=|=|=|=|=|=|=|}
  \multirow{6}{*}{DBpedia} & \multirow{3}{*}{25\%} 
            & seen    & 0.322 & 0.377 & 0.250 & 0.895 & \textbf{0.985} & 0.975 \\
        &   & unseen  & 0.372 & \textbf{0.426} & 0.267 & 0.046 & 0.204 & 0.402 \\
        &   & overall & 0.334 & 0.386 & 0.254 & 0.713 & 0.818 & \textbf{0.852} \\ \cline{2-9}
                        & \multirow{3}{*}{50\%} 
            & seen    & 0.358 & 0.401& 0.202 & 0.960 & \textbf{0.991} & 0.982 \\
        &   & unseen  & 0.304 & \textbf{0.369} & 0.259 & 0.044 & 0.069 & 0.197 \\
        &   & overall & 0.333 & 0.386 & 0.230 & 0.502 & 0.530 & \textbf{0.590} \\
  \hline
  \multirow{6}{*}{20news} & \multirow{3}{*}{25\%} 
            & seen    & 0.205 & 0.279 & 0.263 & 0.614 & \textbf{0.792} & 0.745 \\
        &   & unseen  & 0.201 & \textbf{0.287} & 0.149 & 0.065 & 0.134 & 0.280 \\
        &   & overall & 0.204 & 0.280 & 0.236 & 0.482 & \textbf{0.633} & \textbf{0.633} \\ \cline{2-9}
                        & \multirow{3}{*}{50\%} 
            & seen    & 0.219 & 0.293 & 0.275 & 0.709 & 0.684 & \textbf{0.767} \\
        &   & unseen  & 0.196 & \textbf{0.266} & 0.126 & 0.052 & 0.126 & 0.168 \\
        &   & overall & 0.207 & 0.280 & 0.200 & 0.381 & 0.405 & \textbf{0.469} \\
  \hline
\end{tabular}
\end{table*}

\subsection{Results and Discussion}

\textbf{The evaluation of Phase 1} (coarse-grained classification) checks if each $x_i$ was correctly delivered to the right classifier in Phase 2. 
Table \ref{tab:phase1} shows the performance of Phase 1 with and without augmented data compared with DOC. Considering test documents from seen classes only, our framework
outperformed DOC on both datasets. In addition, the augmented data improved the accuracy of detecting documents from unseen classes clearly and led to higher overall accuracy in every setting. 
Despite no real labelled data from unseen classes, the augmented data generated by topic translation helped Phase 1 better detect documents from unseen classes. Table \ref{tab:augexample} shows some examples of augmented data from the DBpedia dataset. Even if they are not completely understandable, they contain the tone of the target classes. 

\begin{table}[t!]
 \caption{The accuracy of Phase 1 with and without augmented data compared with DOC .} \label{tab:phase1}
\centering
\small
\begin{tabular}{| c | c || c | c | c |} 
  \hline
  \makecell{Dataset \\ Unseen rate} & $y_i$ & DOC & \makecell{Ours \\ w/o aug.} & \makecell{Ours \\ w/ aug.} \\
  \hhline{|=|=|=|=|=|}
  \multirow{3}{*}{\makecell{DBpedia \\ 25\%}} 
            & seen    & 0.980 & \textbf{0.982} & \textbf{0.982} \\
            & unseen  & 0.471 & 0.388 & \textbf{0.536} \\
            & overall & 0.871 & 0.855 & \textbf{0.886} \\ \cline{1-5}
 \multirow{3}{*}{\makecell{DBpedia \\ 50\%}} 
            & seen    & 0.983 & 0.986 & \textbf{0.987} \\
            & unseen  & 0.384 & 0.345 & \textbf{0.512} \\
            & overall & 0.684 & 0.666 & \textbf{0.749} \\
  \hline
  \multirow{3}{*}{\makecell{20news \\ 25\%}} 
            & seen    & 0.800 & \textbf{0.838} & 0.831 \\
            & unseen  & 0.573 & 0.431 &\textbf{0.577} \\
            & overall & 0.745 & 0.754 & \textbf{0.770} \\ \cline{1-5}
  \multirow{3}{*}{\makecell{20news \\ 50\%}} 
            & seen    & 0.824 & \textbf{0.856} & 0.843 \\
            & unseen  & 0.562 & 0.419 & \textbf{0.603} \\
            & overall & 0.694 & 0.639 & \textbf{0.724} \\
  \hline
\end{tabular}
\end{table}

\begin{table}[t!]
 \caption{Examples of augmented data translated from a document of the original class ``Animal'' into two target classes ``Plant'' and ``Athlete''.} \label{tab:augexample}
\centering
\small
\begin{tabular}{| c | m{0.7\linewidth} |} 
  \hline
  \makecell{Animal \\ (Original)} & Mitra perdulca is a species of sea snail a marine gastropod mollusk in the family Mitridae the miters or miter snails. \\  \hline 
  \makecell{Animal \\ $\rightarrow$ \\ Plant} & Arecaceae perdulca is a flowering of port aster a naval mollusk gastropod in the fabaceae Clusiaceae the tiliaceae or rockery amaryllis. \\ \hline
  \makecell{Animal \\ $\rightarrow$ \\ Athlete} & Mira perdulca is a swimmer of sailing sprinter an Olympian limpets gastropod in the basketball Middy the miters or miter skater. \\
  \hline

\end{tabular}
\end{table}

\begin{figure}[t!]
\centering
\includegraphics[width=0.45\textwidth]{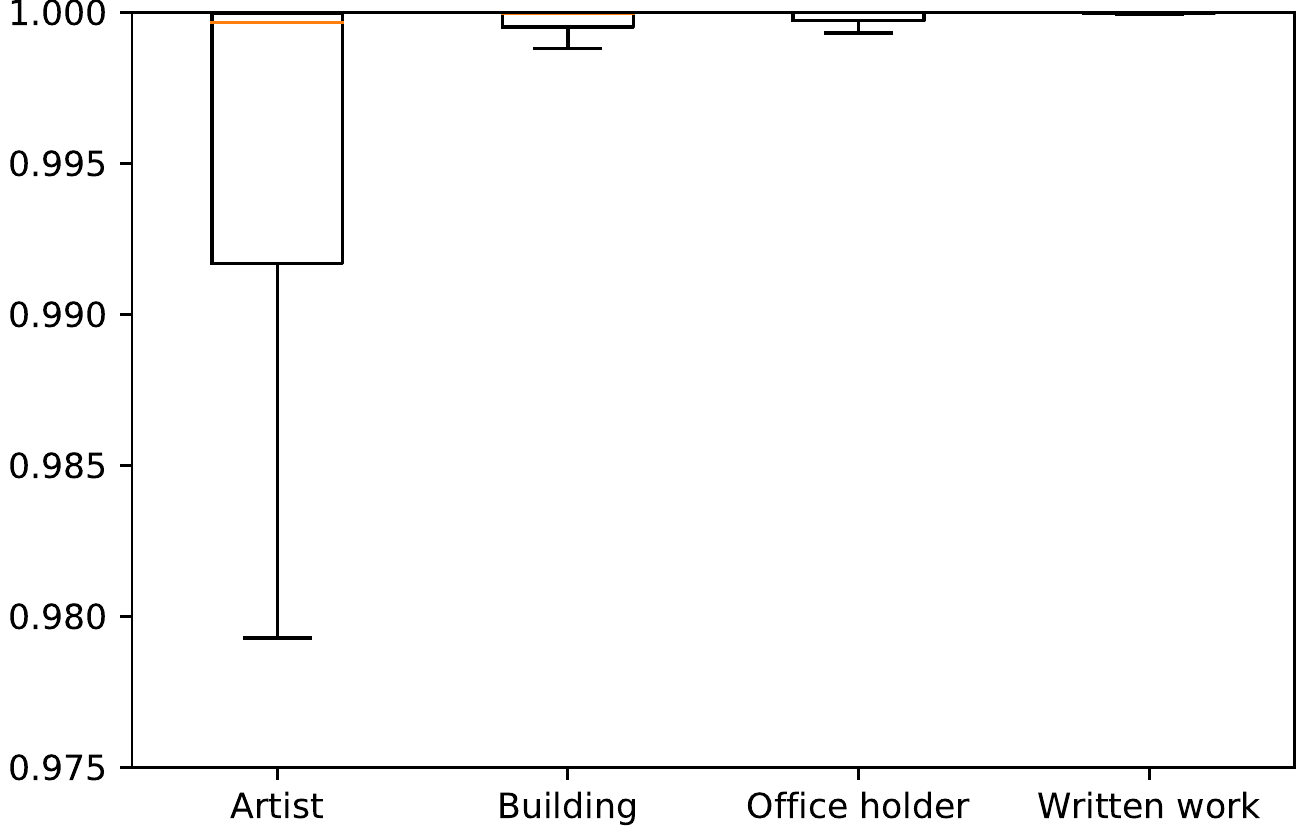}
\caption{The distributions of confidence scores of positive examples from four seen classes of DBpedia in Phase 1. 
}\label{fig:boxplot}
\end{figure}

Although Phase 1 provided confidence scores for all seen classes, we could not use them to predict $\hat{y}_i$ directly since the distribution of scores of positive examples from different CNNs are different. 
Figure \ref{fig:boxplot} shows that 
the distribution of confidence scores of the class ``Artist'' had a noticeably larger variance and was clearly different from the class ``Building''. 
Hence, even if $p(\hat{y}_i=\mbox{``Building''}|x_i) > p(\hat{y}_i=\mbox{``Artist''}|x_i)$, we cannot conclude that $x_i$ is more likely to come from the class ``Building''.
This is why \textbf{a traditional classifier in Phase 2 is necessary}.

Regarding Phase 2, fine-grained classification is in charge of predicting $\hat{y}_i$ and it employs two classifiers which were tested separately. 
Assuming Phase 1 is perfect, the classifiers in Phase 2 should be able to find the right class. The purpose of Table \ref{tab:phase2_seen} is to show that \textbf{the traditional CNN classifier in Phase 2} was highly accurate. 

Besides, given test documents from unseen classes only, the performance of \textbf{the zero-shot classifier in Phase 2} 
is shown in Table \ref{tab:phase2_unseen}. Based on the construction method, $v_{w,c}$ quantified the relatedness between words and the class 
but, unlike $v_w$ and $v_c$, it did not include detailed semantic meaning. Thus, the classifier using $v_{w,c}$ only could not find out the correct unseen class and neither $[v_w; v_{w,c}]$ and $[v_c; v_{w,c}]$ could do.
On the other hand, the combination of $[v_w; v_c]$, which included semantic embeddings of both words and the class label, increased the accuracy of predicting unseen classes clearly. 
However,
the zero-shot classifier fed by the combination of all three types of inputs $[v_w; v_c; v_{w,c}]$
achieved the highest accuracy in all settings. 
It asserts that the integration of semantic knowledge we proposed is an effective means for knowledge transfer from seen to unseen classes
in the zero-shot scenario.

\begin{table}[t!]
 \caption{The accuracy of the traditional classifier in Phase 2 given documents from seen classes only.} \label{tab:phase2_seen}
\centering
\small
\begin{tabular}{| c || c | c | c | c |} 
  \hline
  Dataset & \multicolumn{2}{|c|}{DBpedia} & \multicolumn{2}{|c|}{20news} \\ \hline
  Input $\backslash$ Unseen rate & 50\% & 25\% & 50\% & 25\% \\ \hhline {|=|=|=|=|=|}
  $v_w$ & 0.993 & 0.992 & 0.878 & 0.861 \\ \hline
\end{tabular}
\end{table}

\begin{table}[t!]
 \caption{The accuracy of the zero-shot classifier in Phase 2 given documents from unseen classes only.} \label{tab:phase2_unseen}
\centering
\small
\begin{tabular}{| c || c | c | c | c |} 
  \hline
  Dataset & \multicolumn{2}{|c|}{DBpedia} & \multicolumn{2}{|c|}{20news} \\ \hline
  Inputs $\backslash$ Unseen rate & 50\% & 25\% & 50\% & 25\% \\ \hhline {|=|=|=|=|=|}
  Random guess & 0.143 & 0.333 & 0.100 & 0.200 \\ \hline
  $v_{w,c}$ & 0.154 & 0.443 & 0.104 & 0.210 \\ \hline
  $[v_c; v_{w,c}]$ & 0.163 & 0.400 & 0.099 & 0.215 \\ \hline
  $[v_w; v_{w,c}]$ & 0.266 & 0.460 & 0.122 & 0.307 \\ \hline
  $[v_w; v_c]$ & 0.381 & 0.711 & 0.274 & 0.431 \\ \hline
  $[v_w; v_c; v_{w,c}]$ & \textbf{0.418} & \textbf{0.754} & \textbf{0.302} & \textbf{0.500} \\ \hline
\end{tabular}
\end{table}

Last but most importantly, we compared \textbf{the whole framework} with four baselines as shown in Table \ref{tab:overall}. 
First, the count-based model is a rule-based model so 
it failed to predict documents from seen classes accurately and resulted in unpleasant overall results. 
This was similar to the label similarity approach even though it had higher degree of flexibility. 
Next, the RNN Autoencoder was trained without any supervision since $\hat{y}_i$ was predicted based on the cosine similarity. We believe 
the implicit semantic relatedness between classes caused the failure of the RNN Autoencoder. 
Besides, the CNN+FC and RNN+FC had same inputs and outputs and it was clear that CNN+FC performed better than RNN+FC in the experiment.
However, neither CNN+FC nor RNN+FC was able to transfer the knowledge learned from seen to unseen classes. 
Finally, our two-phase framework has competitive prediction accuracy on unseen classes while maintaining the accuracy on seen classes. This made it achieve the highest overall accuracy on both datasets and both unseen rates. 
In conclusion, by using integrated semantic knowledge, the proposed two-phase framework with data and feature augmentation is a promising step to tackle this challenging zero-shot problem. 

Furthermore, another benefit of the framework is high flexibility. As the modules in Figure \ref{fig:framework} has less coupling to one another, it is flexible to improve or customise each of them. For example, we can deploy an advanced language understanding model, e.g., BERT \cite{devlin2018bert}, as a traditional classifier. Moreover, we may replace ConceptNet with a domain-specific knowledge graph to deal with medical texts. 



\section{Related Work} \label{sec:rel}
\subsection{Zero-shot Text Classification}
There are a few more related works to discuss besides recent approaches we compared with in the experiments (explained in section \ref{subsec:baselines}). 
\citeauthor{2013-Dauphin-ZSLSemanticUtterance} \shortcite{2013-Dauphin-ZSLSemanticUtterance} predicted semantic utterance of texts by mapping class labels and text samples into the same semantic space and classifying each sample to the closest class label. 
\citeauthor{nam2016all} \shortcite{nam2016all} learned the embeddings of classes, documents, and words jointly in the learning stage. Hence, it can perform well in domain-specific classification, but this is possible only with a large amount of training data. 
Overall, most of the previous works exploited semantic relationships between classes and documents via embeddings.  
In contrast, our proposed framework leverages not only the word embeddings but also other semantic knowledge. 
While word embeddings are used to solve analogy for data augmentation in Phase 1, the other semantic knowledge sources (in Figure \ref{fig:semantic}) are integrated into relationship vectors and used as augmented features in Phase 2. Furthermore, our framework does not require any semantic correspondences between seen and unseen classes.



\subsection{Data Augmentation in NLP}
In the face of insufficient data, data augmentation has been widely used to improve generalisation of deep neural networks especially in computer vision \cite{krizhevsky2012imagenet} and multimodality \cite{dong2017i2t2i}, but it is still not a common practice in natural language processing. 
Recent works have explored data augmentation in NLP tasks such as machine translation and text classification \cite{saito2017improving,fadaee2017data,kobayashi2018contextual}, and the algorithms were designed to preserve semantic meaning of an original document by using synonyms \cite{zhang2015text} or adding noises \cite{xie2017data}, for example.
In contrast, our proposed data augmentation technique translates a document from one meaning (its original class) to another meaning (an unseen class) by analogy in order to substitute unavailable labelled data of the unseen class.

\subsection{Feature Augmentation in NLP}
Apart from improving classification accuracy, feature augmentation is also used in domain adaptation to transfer knowledge between a source and a target domain \cite{pan2010survey,fang2018discriminative,chen2018semantic}. 
An early research paper applying feature augmentation in NLP is \citet{hal2007frustratingly} which targeted domain adaptation on sequence labelling tasks. After that, feature augmentation was used in several NLP tasks such as cross-domain sentiment classification \cite{pan2010cross}, multi-domain machine translation \cite{clark2012one}, semantic argument classification \cite{dina2018onfeature}, etc. 
Our work is different from previous works not only that we applied this technique to zero-shot text classification but also that we integrated many types of semantic knowledge to create the augmented features.


\section{Conclusion and Future Work}\label{sec:con}
To tackle zero-shot text classification, we proposed a novel CNN-based two-phase framework together with data augmentation and feature augmentation. 
The experiments show that data augmentation by topic translation improved the accuracy in detecting instances from unseen classes, while feature augmentation enabled knowledge transfer from seen to unseen classes for zero-shot learning.    
Thanks to the framework and the integrated semantic knowledge, our work achieved the highest overall accuracy compared with all the baselines and recent approaches in all settings.
In the future, we plan to extend our framework to do multi-label classification with a larger amount of data, and also study how semantic units defined by linguists can be used in the zero-shot scenario.

\section*{Acknowledgments}
We would like to thank Douglas McIlwraith, Nontawat Charoenphakdee, and three anonymous reviewers for helpful suggestions. 
Jingqing and Piyawat would also like to thank the support from the LexisNexis\raisebox{1ex}{\small{\textregistered}} Risk Solutions HPCC Systems\raisebox{1ex}{\small{\textregistered}} academic program and Anandamahidol Foundation, respectively.

\bibliography{naaclhlt2019}
\bibliographystyle{acl_natbib}

\appendix

\end{document}